\documentclass[conference]{IEEEtran}
\IEEEoverridecommandlockouts
\usepackage{cite}
\usepackage{amsmath,amssymb,amsfonts}

\usepackage[ruled,vlined]{algorithm2e}

\SetKwInput{KwInput}{Input}   
\SetKwInput{KwOutput}{Output} 

\usepackage[utf8]{inputenc} 
\usepackage[T1]{fontenc} 
\usepackage{graphicx} 
\usepackage{booktabs} 
\usepackage{multirow}
\usepackage{array}

\usepackage{graphicx}
\usepackage{textcomp}
\usepackage{xcolor}
\def\BibTeX{{\rm B\kern-.05em{\sc i\kern-.025em b}\kern-.08em
    T\kern-.1667em\lower.7ex\hbox{E}\kern-.125emX}}
\begin{document}

\title{Beyond Dropout: Robust Convolutional Neural Networks Based on Local Feature Masking\\
\thanks{}
}

\author{
	\IEEEauthorblockN{
		Yunpeng Gong\IEEEauthorrefmark{2}, 
		Chuangliang Zhang\IEEEauthorrefmark{2}, 
		Yongjie Hou\IEEEauthorrefmark{1}, 
		Lifei Chen\IEEEauthorrefmark{3}, 
		Min Jiang\IEEEauthorrefmark{2}\IEEEauthorrefmark{2}
	}
	
	\IEEEauthorblockA{\IEEEauthorrefmark{2}School of Informatics, Xiamen University, Xiamen, China}
	\IEEEauthorblockA{\IEEEauthorrefmark{1}School of Electronic Science and Engineering, Xiamen University, Xiamen, China}
	\IEEEauthorblockA{\IEEEauthorrefmark{3}College of Computer and Cyber Security, Fujian Normal University, Fuzhou, China}
	
	\IEEEauthorblockA{
		Email: fmonkey625@gmail.com, 
		\{31520231154325,23120231150268\}@stu.xmu.edu.cn, 
		clfei@fjnu.edu.cn,
		minjiang@xmu.edu.cn
	}
	
	\thanks{\IEEEauthorrefmark{2}\IEEEauthorrefmark{2} Min Jiang is the corresponding author.}
}


\maketitle

\begin{abstract}

In the contemporary of deep learning, where models often grapple with the challenge of simultaneously achieving robustness against adversarial attacks and strong generalization capabilities, this study introduces an innovative Local Feature Masking (LFM) strategy aimed at fortifying the performance of Convolutional Neural Networks (CNNs) on both fronts. During the training phase, we strategically incorporate random feature masking in the shallow layers of CNNs, effectively alleviating overfitting issues, thereby enhancing the model's generalization ability and bolstering its resilience to adversarial attacks. LFM compels the network to adapt by leveraging remaining features to compensate for the absence of certain semantic features, nurturing a more elastic feature learning mechanism. The efficacy of LFM is substantiated through a series of quantitative and qualitative assessments, collectively showcasing a consistent and significant improvement in CNN's generalization ability and resistance against adversarial attacks—a phenomenon not observed in current and prior methodologies. The seamless integration of LFM into established CNN frameworks underscores its potential to advance both generalization and adversarial robustness within the deep learning paradigm. Through comprehensive experiments, including robust person re-identification baseline generalization experiments and adversarial attack experiments, we demonstrate the substantial enhancements offered by LFM in addressing the aforementioned challenges. This contribution represents a noteworthy stride in advancing robust neural network architectures.

\end{abstract}


\section{Introduction}

In the realm of deep learning, the concepts of generalization ability and adversarial robustness stand as two crucial yet often conflicting aspects. The robustness of deep learning models can be comprehensively discussed from the perspectives of the model's generalization ability and adversarial robustness. Although these two aspects overlap, they address different challenges and considerations.

The generalization ability of a model refers to its capacity to maintain performance when exposed to unseen data\cite{1,2,6,7}. In practical applications, models seldom encounter situations identical to their training set, making generalization a key indicator of their utility. The generalization ability significantly depends on the diversity and representativeness of the training data. Deep learning models are particularly prone to overfitting, where they learn features and noise specific to the training data rather than the underlying rules of data generation. Techniques such as weight decay and Dropout serve to enhance a model's generalization ability. Model complexity is also a crucial factor, as overly complex models may perform well on training data but poorly on new data.

Adversarial robustness pertains to a model's resilience when facing deliberately designed minor perturbations, i.e., adversarial attacks\cite{3,4}. Adversarial attacks involve precise modifications to input data to deceive the model into making incorrect predictions. Adversarial training, which introduces adversarial samples into the training set, is a common method to enhance a model's adversarial robustness. Additionally, mechanisms to detect potential adversarial inputs or mitigation strategies such as input preprocessing can be employed to reduce the impact of adversarial attacks.

While generalization ability focuses on the model's capacity to handle normal but unknown data, adversarial robustness concentrates on the model's performance when facing deliberately manufactured perturbations. Improving generalization ability usually involves avoiding overfitting and enhancing data representativeness, whereas enhancing adversarial robustness may require specific training techniques, such as adversarial training. In some cases, improving a model's adversarial robustness can indirectly strengthen its generalization ability.

The robustness issue of deep learning models is multidimensional, involving both the model's generalization ability and adversarial robustness. Enhancing a model's generalization ability necessitates attention to data diversity and representativeness, preventing overfitting, and optimizing model complexity. On the other hand, enhancing adversarial robustness requires attention to the characteristics of adversarial attacks, employing adversarial training, and other mitigation strategies. Both aspects are crucial for building reliable and safe deep learning systems.

In this paper, we introduce a novel regularization technique in the domain of deep learning, termed "Local Feature Masking (LFM)." The primary objective is to achieve a dual enhancement of adversarial robustness and model generalization performance during the training process of deep convolutional neural networks. We empirically demonstrate the effectiveness of LFM in improving both adversarial robustness and model generalization performance, providing a fresh perspective for future neural network model designs.

The main contributions of this paper are summarized as follows:

$\bullet$ We propose a simple yet effective regularization method by introducing the LFM network component. This approach involves random masking of local regions in feature maps at the shallow layers of deep convolutional neural networks during training. The aim is to augment the network's adversarial robustness and generalization capabilities. We conduct a comprehensive analysis of LFM's parameter settings and validate its improvement on the generalization performance of convolutional neural networks through quantitative and qualitative experiments.

$\bullet$ Through simulations of black-box attacks, we showcase that the inherent randomness of LFM significantly enhances the network's adversarial robustness. Compared to existing methods, LFM successfully boosts the model's performance metrics after adversarial attacks, a feat not easily achieved by current regularization techniques. This characteristic endows LFM with a unique advantage in enhancing the robustness of deep learning models.

\section{Related Work}
\subsubsection{Person Re-identification}
Person Re-identification (ReID) is a pivotal task in the field of computer vision, aiming to recognize and track individuals across various time points and camera views within video sequences or images\cite{5,6,8,85,86,87}. This task finds extensive applications in surveillance, video analysis, and intelligent transportation systems. ReID, however, confronts formidable challenges, including variations in pose, changes in lighting conditions, occlusions, and low resolutions. The subtle differences in the appearance of pedestrians further exacerbate the difficulty of distinguishing between different individuals. To address the intricacies of ReID, robust features must be extracted from images or video frames. Convolutional Neural Networks (CNNs) within the domain of deep learning are commonly employed to learn high-level features from images. Feature extraction networks for ReID often incorporate pre-trained CNN architectures such as ResNet and Inception. A key technology in ReID is metric learning, which is essential for measuring the similarity between two pedestrian images. Commonly employed metric learning methods include Euclidean distance and Cosine similarity. Learning an appropriate metric ensures that the feature representations of the same individual are more closely aligned, while those of different individuals are distinctly separated.

\subsubsection{Generalization Ability}
In traditional neural networks, due to the coupling between neurons, the gradient information of one neuron's backpropagation is also influenced by other neurons. This "domino effect" is known as the "complex cooperative adaptation" effect. Dropout \cite{58,59}, as a recognized effective regularization method, randomly sets hidden unit activations to zero with a certain probability during training. This breaks the cooperative adaptation of feature detectors, something that L1, L2, and traditional regularization methods cannot achieve.

Although Dropout has been found to be very effective in regularizing fully connected layers, its effectiveness diminishes when used with convolutional layers. The decrease in effectiveness is primarily attributed to two factors. Firstly, convolutional layers have significantly fewer parameters compared to fully connected layers, thus requiring less regularization. Secondly, neighboring pixels in images and feature maps are interdependent and share much of the same information. If any of them is removed, the information they contain may still be transmitted from the neighboring pixels that remain active. For these reasons, Dropout in convolutional layers only enhances robustness to noisy inputs but lacks the same model-averaging effect observed in fully connected layers.

To enhance the effectiveness of Dropout in convolutional layers, Max-Pooling Dropout \cite{60} proposed a Dropout method for convolutional neural networks. This method applies the Dropout strategy directly to the kernels of the max-pooling layer before performing pooling operations. Spatial Dropout \cite{61} considers applying Dropout to each feature map individually, randomly discarding entire feature maps instead of individual pixels. This effectively circumvents the issue of neighboring pixels passing similar information by randomly removing (setting to zero) some feature maps in each iteration and forcing the network to summarize the remaining feature maps. The drawback of this method is that the number of discarded feature maps is severely limited, and when a significant number of feature maps are discarded, network performance sharply declines because the remaining feature maps cannot reconstruct the lost features from this destructive loss.

Cutout \cite{62} adds noise to input images by randomly applying a rectangular mask as a mask layer during training. This mask layer effectively alters the input images, and the impact of this fixed-position mask layer permeates all feature maps before and after the convolutional neural network. The masked elements are consistently masked, and the remaining elements are consistently retained across all feature maps. This method reduces overfitting and enhances model robustness but lacks efficiency and flexibility. Due to its effectiveness in improving the model's robustness to occluded pedestrians in pedestrian re-identification, it can also effectively enhance model generalization.

\subsubsection{Adversarial Robustness}
Currently, within the academic community, there is no unanimous consensus on adversarial robustness. However, the existing methodologies provide valuable insights, inspiring us to explore new approaches.

In the realm of randomization methods, Xie et al. \cite{72} demonstrated the efficacy of randomly resizing adversarial samples and applying random padding to reduce their adversarial nature. Additionally, evidence suggests that data augmentation techniques during training, such as Gaussian data augmentation \cite{74}, can marginally enhance the neural network's resilience against adversarial attacks.

In the category of network modification methods, Gao et al. \cite{71}, building on the concept of eliminating unnecessary features that could be exploited for generating adversarial samples, identified such features by comparing paired adversarial samples with their clean counterparts. They introduced a masking layer with weights of 0 or 1 as a selector before the classification layer to retain essential features and discard unnecessary ones. Nguyen and Sinha \cite{75} implemented a gradient masking defense by introducing noise to the logit outputs of the network. Kadran et al. \cite{76} modified the output layer of the neural network to enhance its robustness against adversarial attacks. Dhillon et al. \cite{79} proposed Stochastic Activation Pruning (SAP), a method that randomizes the entire network by selectively pruning activations in each layer during forward propagation. This randomization makes the gradients unpredictable, posing challenges in estimating the true gradient accurately. However, the application of SAP leads to a reduction in classification accuracy, indicating its inability to simultaneously enhance both the model's generalization and adversarial robustness.

In the field of ReID~\cite{3,4}, refining model performance demands sophisticated training strategies. Techniques such as warm-up learning rates \cite{56}, label smoothing \cite{57}, and hard sample mining \cite{47,48} have proven instrumental. Additionally, methods like Dropout and its convolutional neural network (CNN) extensions \cite{58,59,60,61,62} complement these strategies.

A pivotal challenge in this domain revolves around striking a delicate balance between model generalization and adversarial robustness. While some methods effectively mitigate overfitting, they often do so at the expense of robustness \cite{63,65,66,71}. Conversely,  approaches emphasizing adversarial robustness may compromise on generalization \cite{63,65,66,71}. This dilemma underscores the intricate nature of simultaneously enhancing both aspects.

This paper introduces local feature masking (LFM), a novel concept inspired by techniques fostering adversarial robustness through randomization \cite{72,73,74} and network modifications \cite{71,75,76,77,78,79}. LFM, a straightforward yet effective regularization method, involves randomly masking local regions in feature maps during the training of CNNs' shallow layers. This strategic approach aims to improve both adversarial robustness and the generalization capabilities of these networks.

\section{Proposed Method}
Drawing inspiration from Dropout and the concept of randomization in existing adversarial defense methods, this study introduces a novel random local feature masking (LFM) strategy. This approach involves the application of localized random masking to feature maps outputted by the shallow layers of the network. Such a strategy diversifies the network's training process and achieves a regularization effect. Moreover, the inherent multiple sources of randomness in the design of the LFM Network notably enhance the model's adversarial robustness. This aspect represents a distinct advantage over Dropout, its extensions in convolutional neural networks, and most existing methods focused on fitting optimization.

The primary motivation behind the LFM Network stems from the issues of overfitting in convolutional neural networks and the vulnerability of deep learning models to adversarial attacks. These issues are common in many other computer vision tasks, such as object recognition, tracking, or human pose estimation. Convolutional neural networks possess powerful representation spaces capable of handling complex learning tasks, and the tens of millions to hundreds of millions of learning parameters in the network provide the necessary representational capacity. However, in cases where the model has too many parameters and too few training samples, the increase in representational capacity also brings the risk of overfitting, leading to a decrease in generalization. Therefore, appropriate regularization is required for good generalization.

Overfitting during training neural networks is characterized by the model having a low loss function and high accuracy on the training data but a high loss function and low accuracy on the test data. Large-scale models can often improve generalization by adding noise to inputs, weights, or gradients during training. A typical example in this regard is Dropout \cite{60}, initially introduced by Hinton and others. However, Dropout and its extensions in convolutional neural networks have certain limitations and have not fully unleashed the potential of the Dropout concept, which is inherent in various key aspects of model training. The proposed Random Masking Layer in this paper operates at the front end of convolutional neural networks. It shares the same advantages as the Dropout method because the changes produced by the masking layer will ultimately affect every layer in the convolutional neural network, including the final fully connected layer. Different regions of masking in different feature maps will suppress the expression of some neurons in a complex way. This is akin to training multiple independent networks on each batch of training data, resulting in an ensemble model of the neural network in the end.

\begin{figure*}[htbp]
\centerline{\includegraphics[width=1\textwidth]{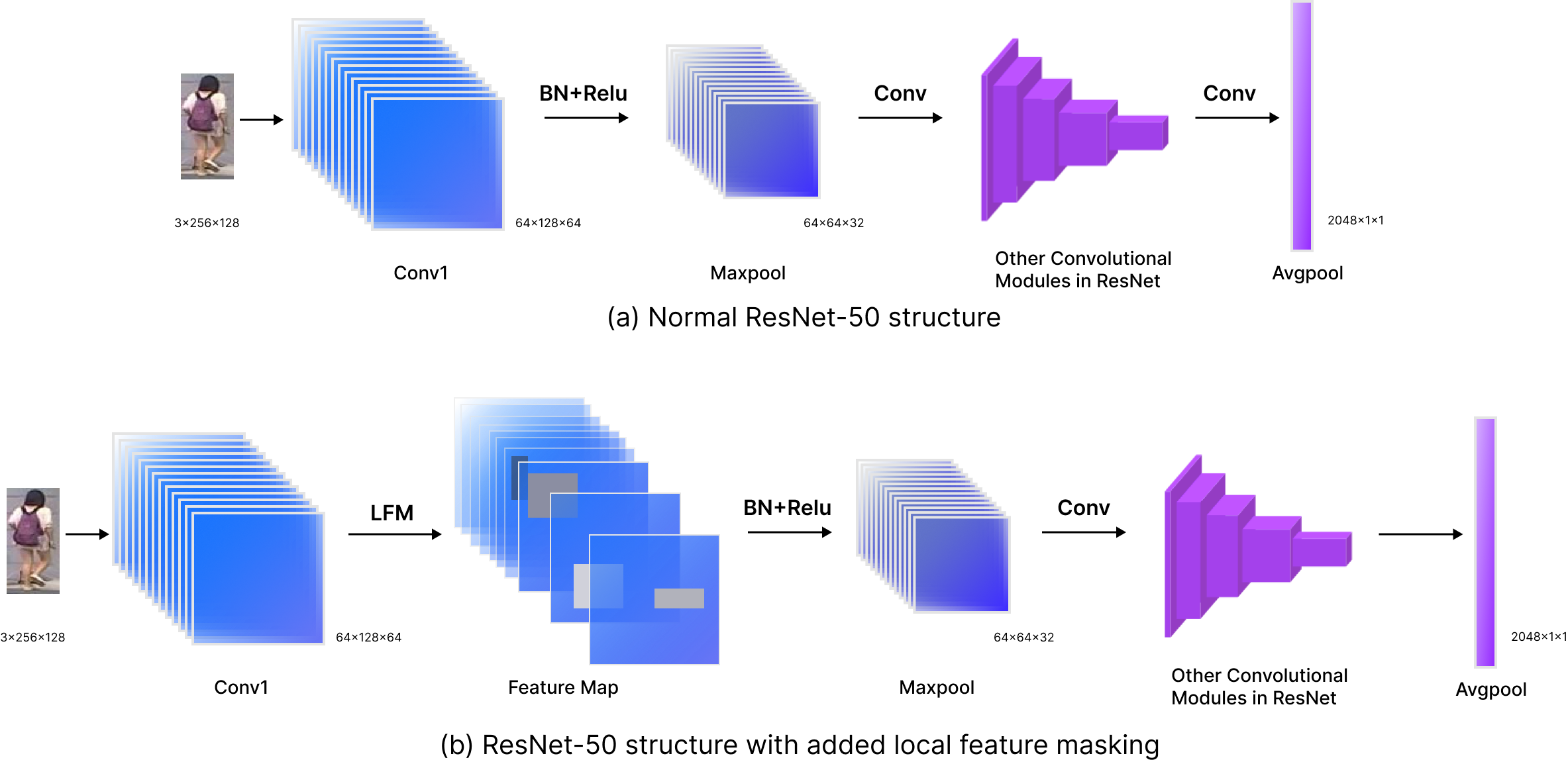}}
\caption{The schematic diagram of the proposed Local Feature Masking in the ResNet network. Fig.(a) illustrates the original ResNet-50 network structure, while Fig.(b) depicts the ResNet-50 network structure with the added Local Feature Masking structure.}
\label{fig}
\end{figure*}

\subsection{Local Feature Masking Algorithm}
As illustrated in Fig. 1, the ResNet-50 backbone network can be segmented into five components. The initial segment encompasses specific operations in the sequence of the conv1 convolutional layer, BN (Batch Normalization), relu activation, and maxpool max-pooling. The subsequent sections, namely conv2\_x, conv3\_x, conv4\_x, and conv5\_x (They are represented in the figure as 'Other Convolutional Modules in ResNet'), each feature residual blocks with analogous structures. Specifically, conv2\_x is composed of 3 Bottlenecks, while the subsequent sections consist of 4, 6, and 3 residual blocks, respectively.

In the initial segment, conv1 conducts convolution using 64 7$\times$7 convolutional kernels with a stride of 2. Following the processing by conv1, the 3-channel RGB image of size 256$\times$128 is initially transformed into a feature map of size 128$\times$64 with 64 channels, encompassing low-level semantic features. As depicted in Fig. 1(b), Local Feature Masking (LFM) applies local random masking to the feature maps generated by conv1, denoted as $I ( I= {I_1, I_2, \ldots, I_n, \ldots, I_{64}})$. Given that the image is normalized before input to the network, a random floating-point number within the range [0, 1] is employed here as the pixel values for the hidden blocks, augmenting the diversity of the masking effect.

The randomness in our LFM manifests in three dimensions. Firstly, the decision to apply feature masking to a sample after its introduction to the neural network is random. Secondly, upon deciding to mask the features of a particular sample, the feature submaps to be masked are chosen randomly. The third dimension involves the random determination of the position and size of the masked region when masking a specific feature submap. It is this triple randomness that endows the feature masking network with the ability to alleviate neural network overfitting while simultaneously enhancing the model's adversarial robustness.

Based on the outlined process, our local feature masking algorithm is formulated as follows:

\begin{algorithm}
	\DontPrintSemicolon
	\caption{Local Feature Masking Procedure}
	
	\KwInput{feature map \( I = \{I_1,I_2,\ldots,I_{64}\} \);\\
		Size of feature map \( I \): \( W \) and \( H \);\\
		Masking probability: \( p \);\\
		Total number of channels in the feature map: \( M \);\\
		Number of channel to be masked: \( N \).\\
		Masking area ratio range \( S_l \) and \( S_h \);\\
		Masking aspect ratio range \( r_1 \) and \( r_2 \).}
	
	\KwOutput{Masked feature map \( I \).}
	
	\textbf{Initialization}: \( p_1 \leftarrow \text{Rand}(0, 1) \).\\
	\If{\( p_1 \geq p \)}{
		\Return \( I \).
	}
	\Else{
		\( \text{Mask\_Channel} = [I_{1}, I_{2}, \ldots, I_{N}] \quad (I_{i} = \text{Rand}(0, M),\; I_{i} \neq I_{j});
		 \) \\
		\For{\( i \) in \( \text{range}(0, N) \)}{
			\( \text{Value} \leftarrow \text{Rand}(0, 1) \), \( S \leftarrow W \times H \);\\
			\While{\text{True}}{
				\( S_e \leftarrow \text{Rand}(S_l, S_h) \times S \);\\
				\( r_e \leftarrow \text{Rand}(r_1, r_2) \);\\
				\( H_e \leftarrow \sqrt{S_e \times r_e} \), \( W_e \leftarrow \sqrt{S_e / r_e} \);\\
				\( x_e \leftarrow \text{Rand}(0, W) \), \( y_e \leftarrow \text{Rand}(0, H) \);\\
				\If{\( x_e + W_e \leq W \) \textbf{and} \( y_e + H_e \leq H \)}{
					\( I_{i}(x_e : x_e + W_e, y_e : y_e + H_e) \leftarrow \text{Value} \);\\
					\textbf{break};
				}
			}
		}
		\Return \( I \).
	}
	
\end{algorithm}

According to Algorithm 1, the input comprises feature maps \(I ( I= \{I_1, I_2, \ldots, I_n, \ldots, I_{64}\})\), and the output is the feature maps processed through local feature masking. Specifically, a random number \(p_1\), in the range \((0,1)\), is obtained using \(\text{Rand}(r_1, r_2)\). If \(p_1 > p\), the feature maps of this sample are not subjected to local feature masking, and the original feature maps are returned. Then, a random number, \(Value\), is generated using \(\text{Rand}(0,1)\) as the pixel value for the \(Mask\_Channel = [I_1, I_i, I_j, \ldots, I_N]\), which are randomly selected. The selected feature submaps are traversed, and for each chosen feature submap \(I_i\), a random rectangular region is generated. Here, \(S_l\) and \(S_h\) represent the minimum and maximum area ratios of the rectangles, and \(S_e = \text{Rand}(S_l, S_h) \times S\) calculates the area size of the random rectangle, constrained within the minimum and maximum ratio. \(r_e\) is a coefficient used to divide the random rectangle area into specific width and height values, determining the shape of the rectangle. It is limited to the \((r_1, r_2)\) interval. Based on empirical knowledge, this paper uses \(S_l = 0.03, S_h = 0.4, r_1 = 0.3\), and \(r_2 = 1/r_1\) as the base settings. \(x_e\) and \(y_e\) are the randomly obtained coordinates of the top-left corner of the target rectangle. If these coordinates would cause the random rectangle to exceed the image boundaries, the area, shape, and position coordinates of the target rectangle are redefined until a suitable rectangle is found. Then, the pixel values of this region are replaced with the pixel value of the randomly generated masking block \(Value\). The same process is applied to all selected feature submap indices, and finally, the feature maps processed through local feature masking are returned. In future work, we will also consider integrating evolutionary computation techniques~\cite{80,81,82,83,84} to extend our method for adaptive optimization.

\begin{figure}[t]
	\centering
	\vspace{1pt} 
	\includegraphics[width=0.5\textwidth, height=0.3\textwidth]{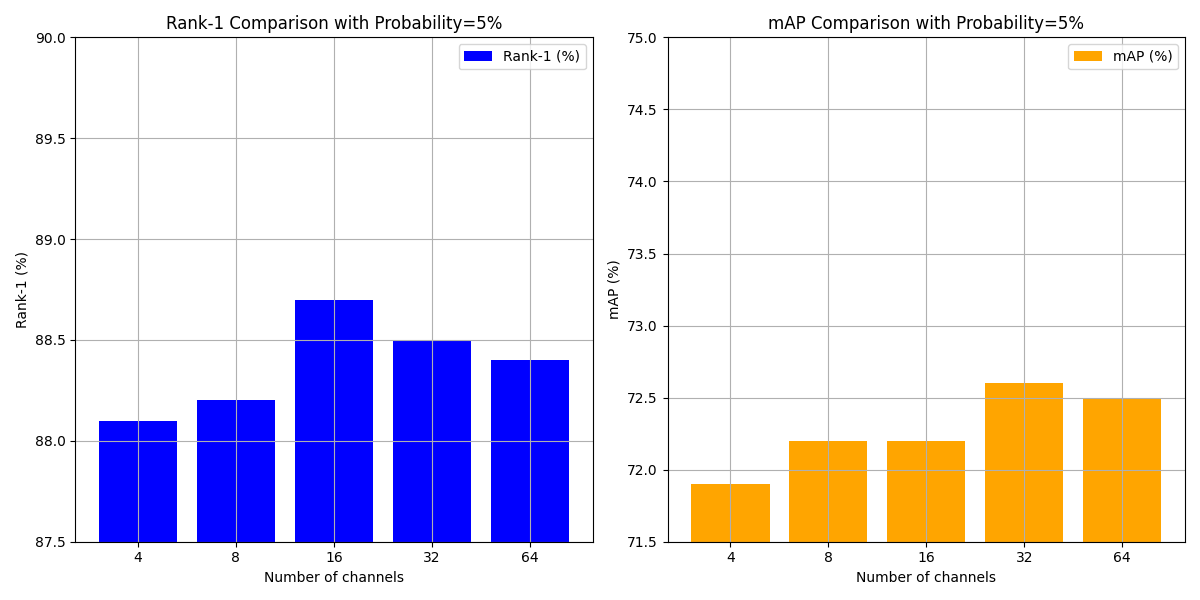}
	\caption{Performance Comparison on the Market1501 Dataset with \( Probability=5\% \).}
	\label{channel}
\end{figure}

\section{Experimental Analysis}
This section will demonstrate the effectiveness of the proposed method through a series of qualitative, comparative experiments, and black-box attack experiments.
\subsection{Datasets and Evaluation Criteria}
The proposed method undergoes evaluation on two person re-identification (ReID) datasets: Market-1501 \cite{9} and DukeMTMC \cite{10}. These datasets are widely acknowledged as the most representative and extensively employed in ReID research. The Market-1501 dataset comprises 12,936 images with 751 identities for training, 19,732 images with 750 identities, and 3,368 query images for testing. DukeMTMC-reID includes 16,522 training images of 702 identities, 2,228 query images of the other 702 identities, and 17,661 gallery images.

Consistent with prior research~\cite{9}, the evaluation utilizes Rank-k precision, Cumulative Matching Characteristics (CMC), and mean Average Precision (mAP) as standard metrics. Rank-1 precision represents the average accuracy of the top-ranked result corresponding to each cross-modality query image. mAP signifies the mean average accuracy, calculated by sorting query results based on similarity. The closer the correct result is to the top of the list, the higher the precision. These metrics collectively offer a comprehensive assessment of the proposed method's performance in comparison to existing works.

As part of a ReID system, re-ranking (reRank)~\cite{11} technology is typically employed to reorganize the initial retrieval results, aiming to more accurately reflect the similarity between images.

\begin{figure}[t]
	\centering
	\includegraphics[width=0.5\textwidth, height=0.3\textwidth]{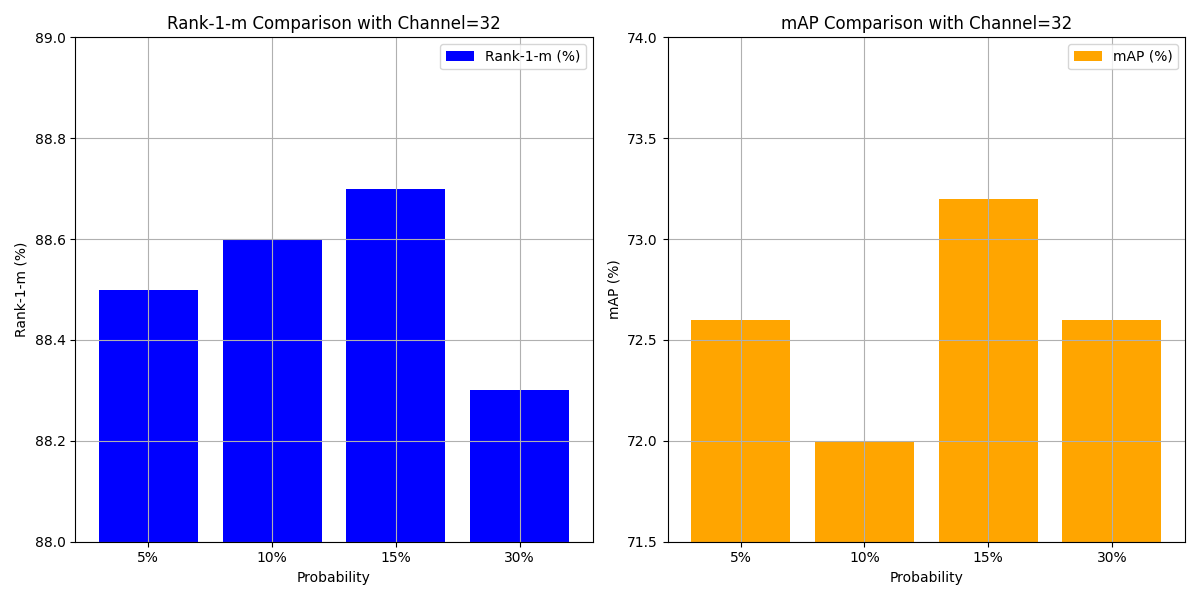}
	\caption{Performance Comparison on the Market1501 Dataset with \( Channel=32 \).}
	\label{p}
\end{figure}

\begin{table*}[]
	\centering
	\vspace{1pt} 
	\resizebox{1\textwidth}{!}{
		\begin{tabular}{lcccc}
			\hline
			\textbf{Method} & { Rank-1(\%) } & {Rank-5(\%)} & {Rank-10(\%)} & {mAP(\%) } \\
			\hline
			ReID Baseline - Dropout & 88.1 & 95.1 & 96.3 & 71.4 \\
			
			ReID Baseline + Dropout & 88.4 & 95.3 & 97.1 & 72.2 \\
			
			ReID Baseline + Dropout + reRank & 90.2 & 97.1 & 98.8 & 84.7 \\
			
			ReID Baseline + LFM (Ours) & 88.7 & 95.3 & 97.7 & 72.4 \\
			
			ReID Baseline + LFM + reRank (Ours) & 90.8 & 97.9 & 98.8 & 85.6 \\
			
			ReID Baseline + LFM + Cutout + Dropout (Ours) & 89.6 & 96.7 & 98.1 & 73.4 \\
			
			ReID Baseline + LFM + Cutout + Dropout + reRank (Ours) & 91.3 & 98.1 & 99.2 & 85.8 \\
			\hline
			Strong Baseline + Dropout & 94.5 & 98.6 & 99.1 & 85.9 \\
			
			Strong Baseline + Dropout + reRank & 95.4 & 99.2 & 99.4 & 94.2 \\
			
			Strong Baseline + LFM (Ours) & 94.8 & 98.8 & 99.2 & 86.6 \\
			
			Strong Baseline + LFM + reRank (Ours) & 95.5 & 99.4 & 99.5 & 94.2 \\
			\hline
		\end{tabular}
	}
	\caption{Performance Comparison of Different Methods on the Market1501 Dataset}
	\label{market1501_methods}
\end{table*}

\begin{table*}[t]
	\centering
	\resizebox{1\textwidth}{!}{
		\begin{tabular}{lcccc}
			\hline
			\textbf{Method} & { Rank-1(\%) } & { Rank-5(\%) } & { Rank-10(\%) } & { mAP(\%) } \\
			\hline
			ReID Baseline + Dropout & 78.7 & 86.2 & 90.5 & 62.3 \\
			
			ReID Baseline + Dropout + reRank & 83.1 & 89.7 & 93.1 & 79.3 \\
			
			ReID Baseline + LFM + Dropout (Ours) & 80.4 & 87.3 & 91.2 & 63.5 \\
			
			ReID Baseline + LFM + Dropout + reRank (Ours) & 84.5 & 90.2 & 93.8 & 79.2 \\
			\hline
			ReID Baseline + Cutout + Dropout & 80.0 & 87.0 & 91.1 & 65.1 \\
			
			ReID Baseline + Cutout + Dropout + reRank & 87.1 & 92.4 & 95.2 & 83.6 \\
			
			ReID Baseline + LFM + Cutout + Dropout (Ours) & 81.4 & 88.4 & 92.0 & 65.1 \\
			
			ReID Baseline + LFM + Cutout  + Dropout + reRank (Ours) & 87.0 & 92.3 & 95.1 & 83.3 \\
			\hline
			Strong Baseline & 86.4 & 91.2 & 94.3 & 76.4 \\
			
			Strong Baseline + reRank & 90.3 & 94.3 & 96.6 & 89.1 \\
			
			Strong Baseline + LFM (Ours) & 87.4 & 92.1 & 95.0 & 77.1 \\
			
			Strong Baseline + LFM + reRank (Ours) & 90.7 & 94.7 & 96.9 & 89.3 \\
			\hline
		\end{tabular}
	}
	\caption{Performance Comparison of Different Methods on the DukeMTMC Dataset}
	\label{dukeMTMC_methods}
\end{table*}

\begin{table*}[]
	\caption{PERFORMANCE COMPARISON OF DIFFERENT Methods UNDER ADVERSARIAL ATTACKS}
	\centering
	\resizebox{1\textwidth}{!}{%
		\begin{tabular}{cccccc|cccc}
			\hline
			\multirow{2}{*}{Dataset} & \multirow{2}{*}{Model} & \multicolumn{4}{c}{Original} & \multicolumn{4}{c}{DMR Attack} \\ \cline{3-10} 
			&        & Rank-1  & Rank-5 & Rank-10   & mAP(\%)    & Rank-1   & Rank-5 & Rank-10  & mAP(\%)    \\ \hline
			\multirow{4}{*}{Market1501} & ResNet                 & 88.0  &95.1  &96.4      & 71.5    & 29.2  & 43.5 & 60.9        & 22.6       \\  
			& ResNet + Dropout   & 88.4   &95.3  &97.1    & 72.2        & 29.7  & 43.8   & 60.1     & 22.3       \\ 
			& ResNet + Dropout + Cutout & 90.2    &96.3  &98.5     & 73.9        & 28.4  & 42.6 & 58.3     & 21.9       \\ 
			& ResNet + LFM (Ours)           & 88.7    &95.8  &97.9     & 72.4        & 34.5 &  44.7 & 63.5     & 24.8       \\ \hline
		\end{tabular}%
	}
	\label{adversarial_attack_performance}
\end{table*}

\subsection{Parameter Configuration and Ablation Experiments}
There are several parameters to be determined in the feature masking layer, one is the number of feature map channels for random feature masking, and the other is the probability of random masking. The remaining parameters are related to the random selection of masking positions in the feature submaps, and these parameters are consistent with the local grayscale transformation\cite{2,1} described in Section of local feature masking algorithm, and will not be reiterated here.

From Fig.~\ref{channel}, random feature masking is performed in the 64 feature maps obtained after the conv1 convolution operation in the first part of the ResNet50 backbone network. Here, with a probability of 5\% for random masking, the number of feature map channels implementing random feature masking is taken as 4, 8, 16, 32, and 64, respectively, and experiments are carried out to determine the optimal parameters for the selection of the number of feature map channels for random feature masking. According to Fig.~\ref{channel}, it can be observed that the performance of the model is optimal when the number of feature map channels for random feature masking, $channels$, is taken to be 32. This means that half of the feature channels in this network layer are involved in random masking, and the masked channels reconstruct the lost information through the other half of the unmasked channels.

Based on the previous experiments, one of the hyperparameters, i.e., the number of feature map channels for random feature masking, was determined to be 32. Next, experiments were conducted with the probability of random masking of 5\%, 10\%, 15\%, and 30\% in order to determine the optimal parameter for the probability of random masking to be taken as the value of the probability of random masking. According to Fig.~\ref{p}, it can be observed that the performance of the model is optimal when the probability of random masking is taken as 15\%. This means that 15\% of the training samples in the process of training the network are trained using the random feature masking network.

\subsection{Comparison experiment}

From TABLE \ref{market1501_methods}, it can be observed by comparing ReID Baseline - Dropout with ReID Baseline on the Market1501 dataset that Dropout helps to improve the model accuracy from 88.1\% to 88.4\% on Rank1, which is a total improvement of 0.3 percentage points, and the mAP is improved from 71.4\% to 72.2\% which improved by a total of 0.8 percentage points. The accuracy of the our LFM help model proposed in this paper is improved from 88.1\% to 88.7\% on Rank1, with a total improvement of 0.6 percentage points; mAP is improved from 71.4\% to 72.4\%, with a total improvement of 1 percentage point. 

The above experiments show that the stochastic masking network proposed in this paper can fully release the potential of the Dropout idea in the convolutional layer, and the performance improvement brought by it even exceeds that of the traditional Dropout method, surpassing 0.3 percentage points in Rank1 index and 0.2 percentage points in mAP. After using the reordering technique re-Rank, it even exceeds the performance by nearly 1 percentage point in the mAP metric.
In addition, this paper also does corresponding experiments on the more advanced benchmark Strong Baseline, and the comparison of Strong Baseline~\cite{12} and Strong Baseline + LFM on mAP metrics shows that the performance improves from 85.9\% to 86.6\%, which is a total of 0.7 percentage points.

In this paper, the same experiments are conducted on the DukeMTMC dataset, and the performance of the methods is consistent with the previous ones as can be seen from the experiments in TABLE~\ref{dukeMTMC_methods}. It is worth noting that the Rank1 accuracy of the model is 78.7\% based on the use of Dropout, which is improved to 80.4\% after combining our LFM proposed in this paper, and the performance is further improved to 81.4\% after combining Cutout on top of both. This strongly suggests that the combination of Cutout, which plays a role in the input layer, our LFM, which plays a role in the front-end of the convolutional neural network, and Dropout, which plays a role in the fully-connected layer at the end of the network, fully unleashes more potential of the idea of Dropout, which is complementary to each other due to the differences in the ways in which they work.

While Cutout does perform well in terms of anti-masking, this approach does not help in terms of improving the adversarial robustness of the model, while the traditional Dropout approach is similarly ineffective in terms of improving adversarial defense. The random masking network proposed in this chapter contains triple randomness, which mitigates network overfitting while also improving the adversarial robustness for the network, and more on adversarial attacks and defenses will be detailed in the next section.

From TABLE \ref{adversarial_attack_performance}, the comparison between the original network ResNet+Dropout and the ResNet+LFM random masking network under DMR black-box attack on the Market1501 dataset, it can be observed that the random masking network effectively improves the model's adversarial robustness, and the model's performance metrics after being attacked on Rank1 improves from 29.7\% to 34.5\%, a total improvement of 4.8 percentage points; mAP improves from 22.6\% to 24.8\%, a total improvement of 2.2 percentage points.

\section{Conclusion}
In this paper, we introduced a novel regularization method, named Local Feature Masking (LFM), to enhance the adversarial robustness and generalization performance of deep convolutional neural networks. By introducing random masking in specific regions during the training process, LFM successfully addresses the challenge of simultaneously improving the adversarial robustness and generalization performance, overcoming the dilemma faced by existing regularization methods.

Through extensive parameter analysis and both quantitative and qualitative experiments, we demonstrated the effectiveness of LFM, showcasing its outstanding performance in enhancing the generalization capabilities of neural networks. Through black-box attack simulations, we illustrated that the inherent randomness of LFM significantly boosts the network's adversarial robustness, effectively tackling the challenges posed by adversarial attacks.

In summary, LFM provides a simple yet powerful regularization tool for deep learning model design, bridging the gap between model generalization and adversarial robustness. Future research could explore extending LFM to other network architectures and tasks to further broaden its applicability. We believe the introduction of LFM will have a profound impact on the deep learning community, laying the foundation for constructing more robust and efficient neural networks.

\section*{Acknowledgment}

This work was partly supported by the National Natural Science Foundation of China under Grant No. 62276222

\vspace{12pt}
\color{red}

\end{document}